\documentclass{bmvc2k}

\title{Improving Personalized Image Generation through Social Context Feedback}

\addauthor{Parul Gupta}{parul@monash.edu}{1}
\addauthor{Abhinav Dhall}{abhinav.dhall@monash.edu}{1}
\addauthor{Thanh-Toan Do}{toan.do@monash.edu}{1}

\addinstitution{
 Monash University\\
 Australia
}

\runninghead{Gupta et al.}{Social Context Personalization}

\usepackage{multirow}
\usepackage{adjustbox}
\usepackage{wrapfig}
\usepackage{booktabs}

\begin{document}

\maketitle
\begin{abstract}
\noindent Personalized image generation, where reference images of one or more subjects are used to generate their image according to a scene description, has gathered significant interest in the community. However, such generated images suffer from three major limitations -- complex activities, such as \textit{$<$man, pushing, motorcycle$>$} are not generated properly with incorrect human poses, reference human identities are not preserved, and generated human gaze patterns are unnatural/inconsistent with the scene description. In this work, we propose to overcome these shortcomings through feedback-based fine-tuning of existing personalized generation methods, wherein, state-of-art detectors of pose, human-object-interaction, human facial recognition and human gaze-point estimation are used to refine the diffusion model. We also propose timestep-based inculcation of different feedback modules, depending upon whether the signal is low-level (such as human pose), or high-level (such as gaze point). The images generated in this manner show an improvement in the generated interactions, facial identities and image quality over three benchmark datasets.
\end{abstract}    
\section{Introduction}
\label{sec:intro}
Recent advancement in generative AI has led to the emergence of several new applications that are now feasible through large-scale diffusion models, trained on internet-scale data. One such application is \textit{personalizing} the generated media content using reference images of objects and/or people. Multiple approaches for subject-driven image generation involve inference-time fine-tuning of a base model~\cite{Ruiz_2023_CVPR, gal2022textual, Kumari_2023_CVPR, Han_2023_ICCV, NEURIPS2023_3340ee1e}, which is both time and compute intensive. Therefore, efforts have been made to develop techniques that do not require fine-tuning for each new generation~\cite{Wei_2023_ICCV, 10657619}. Further improvements have been proposed by addressing the subject-identity mixing and subject-neglect/subject-dominance problems for multi-subject generation~\cite{osti_10543865, wang2025msdiffusion}.
Despite all these refinements, existing methods often fail to showcase the referenced subjects faithfully when the text prompts become more complex, for example, when the activity described in the prompt is more evolved, (also shown in Figure \ref{fig:comparison} later). This limits the utility of such approaches in social context, where humans interact with the objects in their surroundings in a non-trivial manner. To tackle with this issue, we propose utilizing state-of-the-art detectors of human pose, human-object interaction, identity, and gaze to guide the personalization process. Specifically, these detectors provide feedback to the diffusion model \textit{while it is being trained}, thereby enriching it with advanced activity/interaction and gaze-pattern generation capabilities that are inline with both the input text prompt as well as the referenced subject images.
\begin{wrapfigure}{l}{0.5\linewidth}
\centering
    \includegraphics[width=\linewidth]{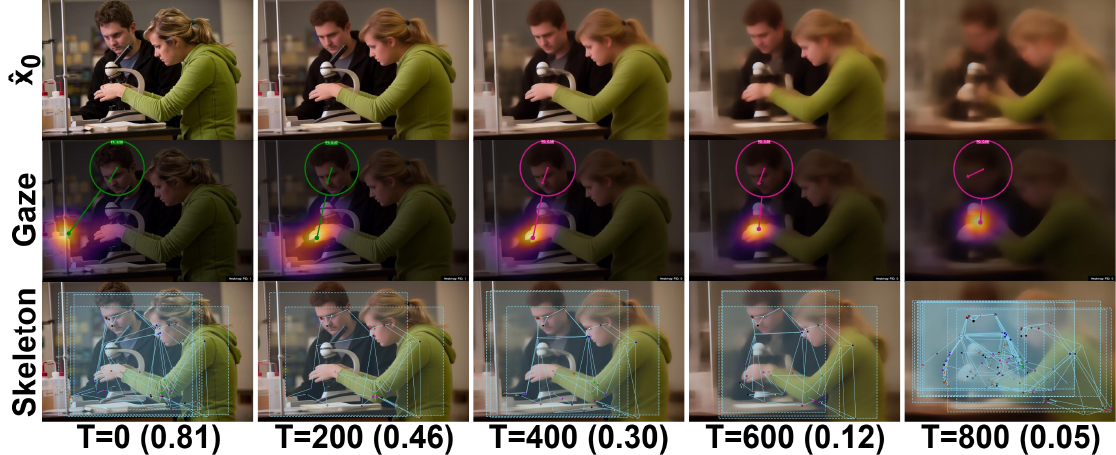}
  \caption{While denoising, diffusion models reconstruct coarse features at higher timesteps and refine finer details as denoising progresses. Consequently, the estimated gaze and facial embeddings diverge earlier (at $t \approx 200$ and $t \approx 400$, respectively) than the skeleton keypoints ($t \approx 800$). Values in brackets denote cosine similarity with the ground-truth facial embedding.}
  \label{fig:analysis}
  \vspace{-\baselineskip}
\end{wrapfigure}
An observation to be noted here is that human pose and human-object interaction are both coarser signals as compared to human identity and human gaze. Several prior works~\cite{Qian_2024_CVPR, avrahami2023blendedlatent} have shown that diffusion models learn high-level (coarse) details at higher noise levels and low-level (fine-grained) information at lower noise levels. Thus, while denoising any given training sample (starting from pure Gaussian noise), the model recovers the human pose and human-object interaction first, and the human identity and gaze are generated later. This is also demonstrated in Figure \ref{fig:analysis} which shows the image recovered ($\mathbf{\hat{x}}_{0}$) at different timesteps using a Stable Diffusion-V$1.5$ model. The detected gaze (using Sharingan~\cite{Tafasca_2024_CVPR}) becomes erroneous at timestep $t=200$, while the cosine similarity between the man's facial identity embeddings (using ArcFace~\cite{Deng_2019_CVPR}) in the original and recovered images drops to $0.30$ at $t=400$ (the minimum cosine similarity should be $0.32$ to be classified as the same person). The detected skeleton-keypoints (using X-Pose~\cite{X_Pose_2024}), on the other hand, are accurate until much later, and become error-prone at $t=800$. Given this information, penalizing the model for incorrect human identity/gaze at higher noise levels would rather destabilize the training. This leads us to segregate the range of noise-levels (or timesteps) where each individual feedback signal is applied. Our feedback-based learning approach can also be applied to fine-tune any existing personalized generation framework. Thus, we summarize our main contributions as follows -- \textbf{(1)} We introduce an approach that improves subject-driven image generation \textit{in social context} by making it pose, interaction, identity and gaze-aware, while maintaining fidelity to the text prompt. \textbf{(2)} We leverage state-of-the-art detectors of different social cues (pose, interaction, identity and gaze) that act as feedback modules \textit{during the training} of the diffusion model to improve the generation quality. \textbf{(3)} The inculcation of feedback from different social cues is timestep-aware, depending upon the coarseness of the signal, that further stabilizes the training.


\section{Related Work}
\label{sec:rel_work}
\textbf{Subject-driven Image Generation} approaches can be mainly classified into four categories depending upon whether only a single or multiple reference subject(s) can be generated; and whether inference-time fine-tuning is required or not. Dreambooth~\cite{Ruiz_2023_CVPR} embeds the subject into the Diffusion model weights, Textual Inversion~\cite{gal2022textual} learns a textual encoding corresponding to the reference subject, MagiCapture~\cite{hyung2023magicapture} learns to generate a reference human subject in a reference style by learning their textual embeddings while masking the denoising objective to focus only on the foreground region for the subject and only on the background region for the style. Custom-Diffusion~\cite{Kumari_2023_CVPR} and SVDiff~\cite{Han_2023_ICCV} provide further improvement by allowing joint learning of multiple concepts and reducing the number of parameters needing fine-tuning. Mix-of-Show~\cite{NEURIPS2023_3340ee1e} learns different embedding-decomposed LoRAs (ED-LoRAs) for different subjects and then combines them using gradient fusion. IP-Adapter~\cite{ye2023ipadaptertextcompatibleimage}, ELITE~\cite{Wei_2023_ICCV} and InstantBooth~\cite{10657619} integrate the global and patch-wise image features of the single reference subject into the UNet by introducing additional cross-attention layers, and do not require inference-time finetuning. More recent works, such as FastComposer~\cite{osti_10543865}, SSR-Encoder~\cite{Zhang_2024_CVPR}, Subject-Diffusion~\cite{Ma_2024_SubjectDiffusion} and MIP-Adapter~\cite{huang2024resolvingmulticonditionconfusionfinetuningfree} enable multi-concept composition without inference-time finetuning. FastComposer achieves this by optimizing the cross-attention maps corresponding to the subject-tokens in the conditional space, to match them with the respective binary segmentation maps of the subjects. SSR-encoder introduces additional cross-attention based alignment module between reference subject texts and images, and MIP Adapter does weighted merging of the reference images' features with the target image's latent features.\\
\noindent Despite all these advancements, as soon as the text prompt becomes more complex, such as including any non-trivial activities, the generated image loses its quality. This is the limitation our feedback based diffusion framework aims to address. Outside the personalization domain, Interact-Diffusion~\cite{Hoe_2024_CVPR} improves the interaction-generation capability of Diffusion Models, by taking the reference subject texts, their bounding boxes and their interaction labels as input. SA-HOI~\cite{xusemantic} proposes an iterative refinement based sampling strategy for generating human-object interactions using text prompts. Similarly, TextGaze~\cite{Wang_TextGaze} proposes a diffusion model based framework to generate head poses and gaze-patterns using textual descriptions of the poses and gaze directions.

\section{Method}
\label{sec:Method}
\begin{figure}[t]
    \centering
    \includegraphics[width=\linewidth]{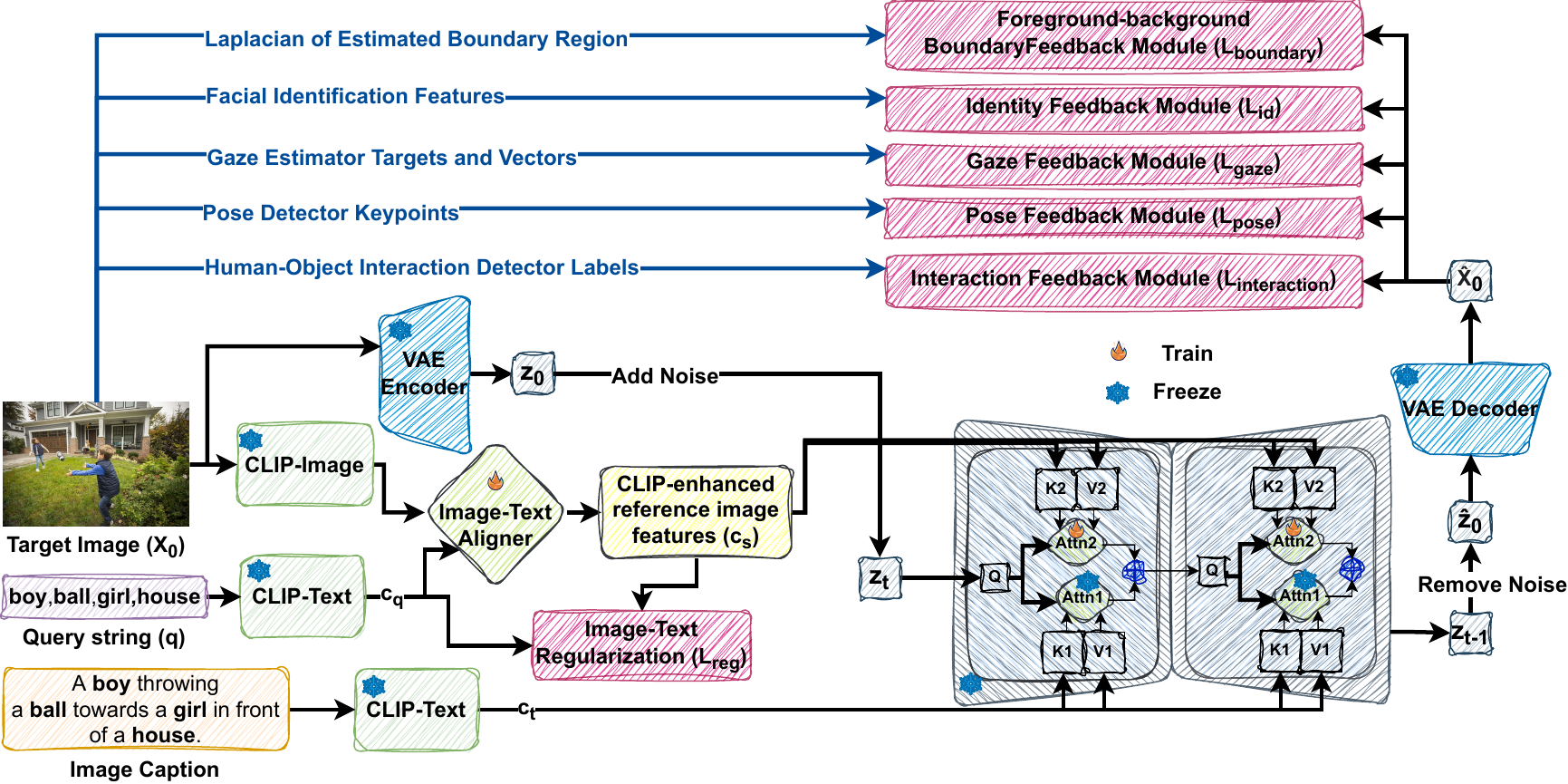}
    \caption{\textbf{Overview of Feedback based Diffusion finetuning: } During training, the input image for the reference subjects is same as the target (to be generated) image $\mathbf{x}_0$. At each timestep $t$, text-aligned image features $\bar{c}_s$ are extracted by aligning the CLIP-Image features of the input image with the CLIP-Text features of the query string ($q$) containing the reference subjects. These aligned features are then ingested into UNet through additional cross attention layers, whose output is added to the output of the text (caption)-based cross-attention layers. Next, the approximate $\mathbf{\hat{x}}_0$ is computed by passing the predicted noise $\mathbf{\epsilon}_\theta$ through the VAE Decoder. Pretrained detectors compute signals (boundary gradients, identity, gaze, pose, interaction) on $\hat{\mathbf{x}}_0$, which are matched against $\mathbf{x}_0$ using respective losses ($\mathcal{L}_{\text{boundary}}, \mathcal{L}_{\text{id}}, \mathcal{L}_{\text{gaze}}, \mathcal{L}_{\text{pose}}, \mathcal{L}_{\text{interaction}}$), along with the standard denoising loss ($\mathcal{L}_{\text{denoise}}$). A regularization loss ($\mathcal{L}_{\text{reg}}$) aligns $\bar{c}_s$ with CLIP-Text features $c_q$.}
    \vspace{-\baselineskip}
    \label{fig:feedback_pipeline}
\end{figure}
We adopt the framework of SSR-Encoder to improve the gaze-fidelity, identity preservation and interaction generation in diffusion models. Given a reference image (containing the referenced subjects; this is the same as the target, or to be generated image while training) $\mathbf{x}_{0}$, a query string (which contains the references subjects' class names, separated by commas) $q$ and a target image caption, SSR-Encoder first extracts multi-scale CLIP image features from the reference image and aligns them with the CLIP text embeddings of the query string ($c_q$) using Image-Text Aligner (a layer based upon cross-attention mechanism). Then these aligned image-text features (denoted by $c_s$), along with the image caption embeddings $c_t$, are injected through separate cross-attention layers into the UNet model, which learns to denoise the target image.\\
The learning is guided by two loss functions, the first is the denoising loss $\mathcal{L}_{\text{denoise}}$, which is the mean squared error between the true noise and the predicted noise, while the other is the regularization loss $\mathcal{L}_{\text{reg}}$ to enhance the cosine similarity between the CLIP query string embeddings ($c_q$) and the average of the aligned image-text features ($\bar{c_s}$). At any denoising timestep $t$, the estimated $\mathbf{\hat{x}}_{0}$ can be obtained as--
\begin{align}
    \mathbf{\hat{z}}_{0}= \left(\mathbf{z}_t-\sqrt{1-\bar\alpha_t}\mathbf{\epsilon}_\theta\left(\mathbf{z}_t\right)\right)/\sqrt{\bar\alpha_t}; \quad\quad \mathbf{\hat{x}}_{0}=\text{VAE-Decoder}\left(\mathbf{\hat{z}}_{0}\right)
\end{align}
\noindent To improve the denoising process, we introduce the following additional supervision signals:\\
\noindent\textbf{A. Foreground-background Boundary Region Feedback ($\mathcal{L}_{\text{boundary}}$):} To improve the coherence between the generated foreground,i.e. the referenced subjects and the rest of the image (background); and overcome the copy-paste artifacts, we enforce that the gradients in the boundary region (between the subjects and the background) should be similar between the target image and the estimated $\mathbf{\hat{x}}_{0}$ at each denoising step.  Therefore, we first extract the boundary region by performing a morphological gradient operation on the binarized segmentation map of the target image $\mathbf{x}_{0}$ (where pixels corresponding to the foreground subjects are $1$). Then we obtain the gradients by performing convolution (of both the ground truth $\mathbf{x}_{0}$ and the estimated $\mathbf{\hat{x}}_{0}$) with a Laplacian kernel.
\begin{equation}
\begin{split}
\mathcal{L}_{\text{boundary}}=\lVert\text{Laplacian}\left(\mathbf{x}_0\right)*\text{Boundary Map} - \text{Laplacian}\left(\mathbf{\hat{x}}_0\right)*\text{Boundary Map}\rVert^2
\end{split}
\end{equation}
\noindent\textbf{B. Facial Identity Feedback ($\mathcal{L}_{\text{id}}$):} To improve the similarity between the reference and generated human faces, we use a pretrained (frozen) facial recognition model ArcFace~\cite{Deng_2019_CVPR} to extract the facial identity embeddings from the valid reference subjects (having a human face). Next, we use the spatial locations of the valid reference subjects in the ground truth $\mathbf{x}_{0}$, and extract the identity embeddings from these locations in the estimated $\mathbf{\hat{x}}_{0}$.
\begin{equation}
\begin{split} 
\mathcal{L}_{\text{id}}=\lVert\text{ArcFace}\left(\mathbf{x}_0\left[\text{face bbox}\right]\right) - \text{ArcFace}\left(\mathbf{\hat{x}}_0\left[\text{face bbox}\right]\right)\rVert^2
\end{split}
\end{equation}
\noindent\textbf{C. Gaze Feedback ($\mathcal{L}_{\text{gaze}}$):} To ensure that the gaze pattern in the generated images is consistent with action/activity described in the text caption, we use a pretrained (frozen) gaze prediction model Sharingan~\cite{Tafasca_2024_CVPR} to detect the gaze targets and the gaze vectors (directed from person's head center to the gaze target) in the ground truth image $\mathbf{x}_{0}$. Then we enforce that the gaze targets and gaze vectors in the estimated $\mathbf{\hat{x}}_{0}$ should match with the ground truth.
\begin{equation}
\begin{split}
\mathcal{L}_{\text{gaze}}=\lVert\left[\text{gt}\right]_{\text{Ground Truth}}- \text{Sharingan}\left(\mathbf{\hat{x}}_0\right)\left[\text{gt}\right]\rVert^2 + 1-\text{cos}\left(\left[\text{gv}\right]_{\text{Ground Truth}}, \text{Sharingan}\left(\mathbf{\hat{x}}_0\right)\left[\text{gv}\right]\right)
\end{split}
\end{equation}
where cos refers to the cosine similarity function, gt denotes the gaze target and gv denotes the gaze vector.\\
\noindent\textbf{D. Pose Feedback ($\mathcal{L}_{\text{pose}}$):} To match the pose of the generated objects with the ground truth target image, we use a pretrained (frozen) generic pose (keypoint) detection model X-Pose~\cite{X_Pose_2024} to obtain the (normalized) keypoint coordinates of reference subjects in both $\mathbf{x}_{0}$ and $\mathbf{\hat{x}}_{0}$ and calculate the mean squared error.
\begin{equation}
\begin{split}
\mathcal{L}_{\text{pose}}=\lVert\text{X-Pose}\left(\mathbf{x}_0\left[\text{subject bbox}\right]\right) - \text{X-Pose}\left(\mathbf{\hat{x}}_0\left[\text{subject bbox}\right]\right)\rVert^2
\end{split}
\end{equation}
\noindent\textbf{E. Interaction Feedback ($\mathcal{L}_{\text{interaction}}$):} To ensure that the generated interaction between the reference subjects (or between a reference subject and some subject in the background) is consistent with the action described in the text caption, we use a pretrained (frozen) human-object interaction (HOI) detector CMMP~\cite{ting2024CMMP} and match the labels of the human-object interaction classes in the ground truth $\mathbf{x}_{0}$ with the logits predicted in the estimated $\mathbf{\hat{x}}_{0}$ by applying Focal loss~\cite{focal_loss_2017}.
\begin{equation}
\mathcal{L}_{\text{interaction}}=\text{Focal loss}\left(\text{HOI label}_{\text{Ground Truth}}, \text{CMMP}\left(\mathbf{\hat{x}}_0\right)\right)
\end{equation}
Thus, the overall loss while finetuning the diffusion model becomes
\begin{equation}
\begin{split}
\mathcal{L}=&\mathcal{L}_{\text{denoise}}+\lambda_{\text{reg}}\mathcal{L}_{\text{reg}}+\lambda_{\text{boundary}}\mathcal{L}_{\text{boundary}}+\lambda_{\text{id}}\mathcal{L}_{\text{id}}+\lambda_{\text{gaze}}\mathcal{L}_{\text{gaze}}+\lambda_{\text{pose}}\mathcal{L}_{\text{pose}}\\&+\lambda_{\text{interaction}}\mathcal{L}_{\text{interaction}}
\end{split}
\end{equation}
\begin{figure}
\begin{subfigure}
    \centering
    \includegraphics[width=0.65\linewidth]{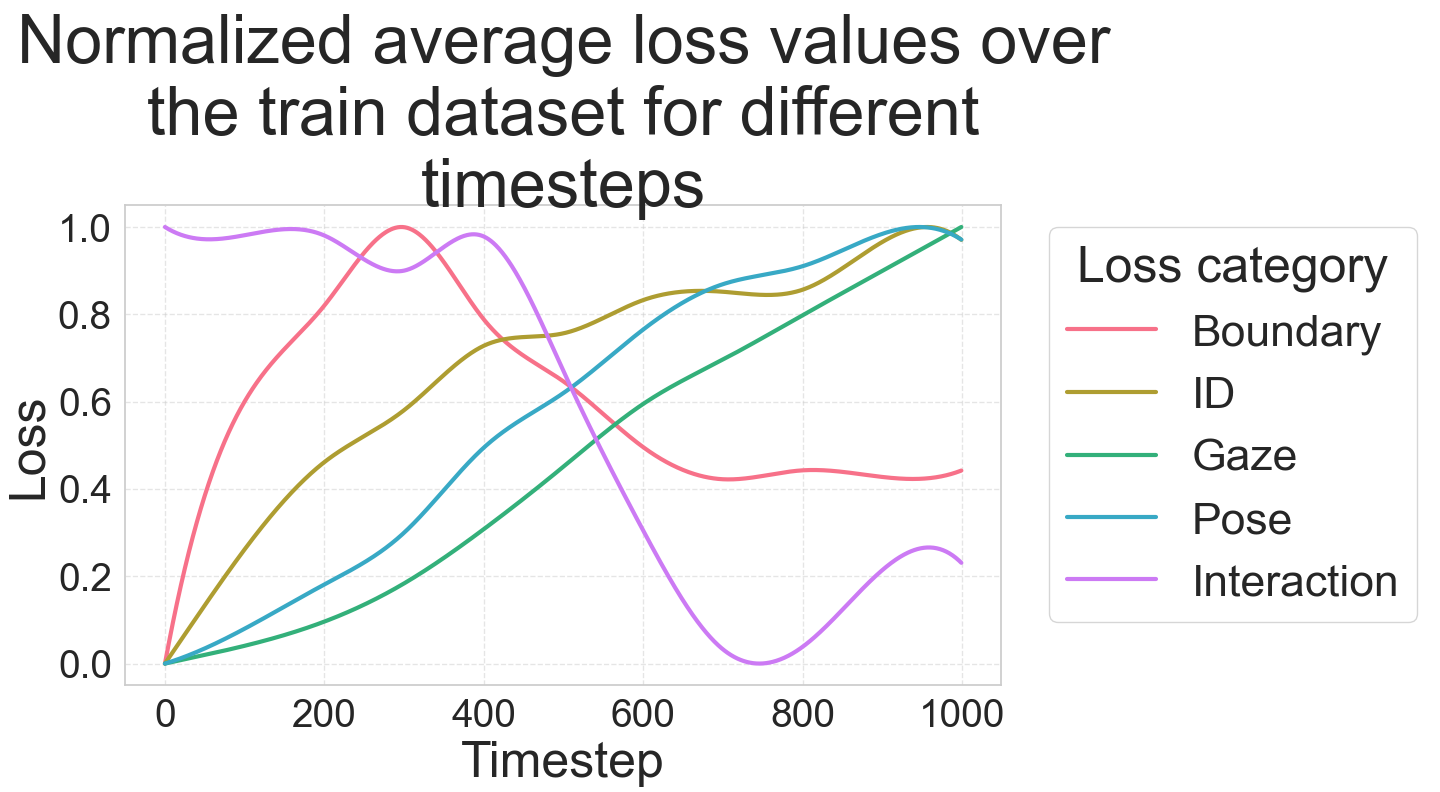}
\end{subfigure}
\begin{subfigure}
    \centering
    \includegraphics[width=0.33\linewidth]{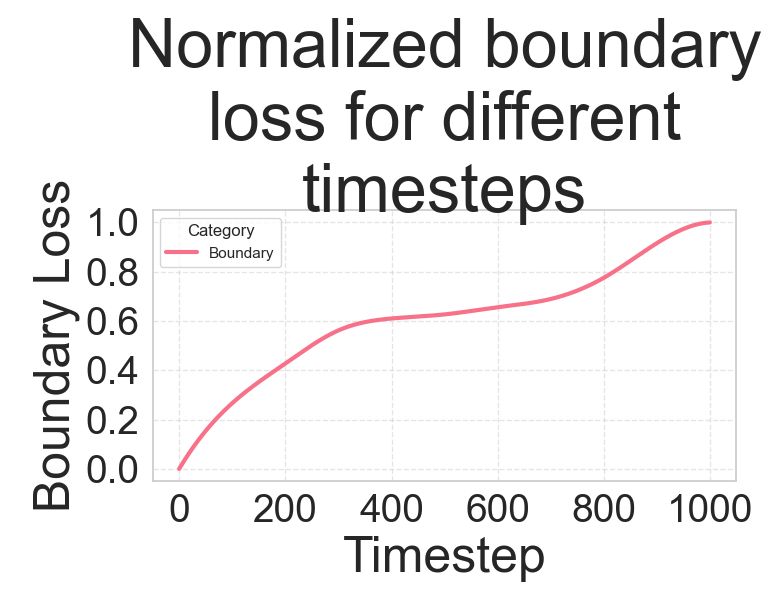}
\end{subfigure}
  \caption{(left) The average loss values over the HICO-DET train split for $\mathcal{L}_{\text{boundary}}, \mathcal{L}_{\text{id}}, \mathcal{L}_{\text{pose}}, \mathcal{L}_{\text{interaction}}$ and over the Gazefollow train split for $\mathcal{L}_{\text{gaze}}$, at different timesteps, calculated using pretrained SSR-Encoder checkpoint, normalized between 0 and 1. (right) The boundary loss curve after the finer details of an image are removed by passing a Gaussian filter.}
  \label{fig:tstep_loss}
\end{figure}
In all our experiments, $\lambda_{\text{reg}}=\lambda_{\text{boundary}}=\lambda_{\text{id}}=\lambda_{\text{gaze}}=\lambda_{\text{pose}}=\lambda_{\text{interaction}}=0.01$, unless otherwise specified.\\
\noindent\textbf{F. Timestep-dependent supervision:}
As discussed in Section \ref{sec:intro}, since gaze, facial identity, interaction and pose have varying levels of granularity, and the denoising process recovers only the coarse details at higher noise levels, we apply these supervision signals only within specific timestep ranges, determined through manual inspection: $\mathcal{L}_{\text{gaze}}$ for $t\in[0,200]$, $\mathcal{L}_{\text{id}}$ for $t\in[0,400]$, whereas $\mathcal{L}_{\text{interaction}}$ for $t\in[0,500]$ and $\mathcal{L}_{\text{pose}}$ for $t\in[0,700]$. This is further supported by the graph in Figure \ref{fig:tstep_loss} (left), which shows the normalized average loss values over the train split at different timesteps. The curves for $\mathcal{L}_{\text{id}}, \mathcal{L}_{\text{gaze}}$ and $\mathcal{L}_{\text{pose}}$, which are mean squared error losses, increase steadily as the respective latent embeddings are further from the real ones. However, there is a sudden decline in $\mathcal{L}_{\text{interaction}}$ (focal loss) around $t=500$, making the detector's predictions unreliable after this timestep. This happens because the interaction loss is calculated only when the predicted object class has a positive prior probability over the correct verb class, and as the noise level increases, the object classes being predicted become more and more erroneous, leading to the interaction loss dropping to zero. Counter-intuitively, $\mathcal{L}_{\text{boundary}}$ also reduces after $t\approx300$, even though the images at higher timesteps are further from the original image as per human perception. This shows that when the finer details are emerging during denoising, the boundary region gradients initially tend to be irregular ($t\in[300,700]$), and later come closer to the original image ($t\in[0,300]$). To verify this, we calculate the boundary loss versus timestep curve for the image shown in Figure~\ref{fig:analysis} after removing its fine details by passing a Gaussian filter of radius 1 over it. The resulting loss curve is increasing (Figure~\ref{fig:tstep_loss} (right)), demonstrating that the counter-intuitive boundary loss is a result of its sensitivity towards the finer details in an image.


\section{Experiments}
\label{sec:experiments}
\subsection{Experimental Setup}
\noindent\textbf{Training Data} We use three benchmark datasets in our experiments-- HICO-DET dataset~\cite{chao:wacv2018} for human-object interaction, GazeFollow dataset~\cite{nips15_recasens} for gaze-pattern and Concept101 dataset~\cite{Kumari_2023_CVPR} for general multi-concept personalization. The train sets of HICO-DET and GazeFollow are used for fine-tuning, while their test sets along with Concept101 are used for evaluation. For both HICO-DET and GazeFollow, we employ BLIP-2~\cite{blip_li_2023} to obtain image captions, and then use the spaCy~\cite{spacy_2020} library to extract entities (i.e. nouns). Next, the object-detection Grounding-DINO~\cite{grounding_dino} model gives the bounding boxes corresponding to the entities, following which, SAM~\cite{kirillov2023segany} predicts the segmentation masks for each of the entities in the image. We also obtain the facial identity embeddings using ArcFace~\cite{Deng_2019_CVPR} and the skeleton-keypoint coordinates using X-Pose~\cite{X_Pose_2024} for each of the entities. For interaction and gaze training, we only use the ground truth annotations available in HICO-DET and GazeFollow datasets respectively. Each interaction class refers to a unique combination of $<$verb, object$>$ labels (subject is always a human). Finally, our processed HICO-DET has $37,600$ images in the train set and $9,546$ images in the test set. The GazeFollow dataset has $86,900$ images in the train set and $4,756$ images in the test set.

\noindent\textbf{Implementation Details} We start with the pretrained checkpoint of SSR-Encoder and use the same image preprocessing steps (shortest side resized to $512$, followed by random crop to get $512\times512$ resolution) while finetuning. The input to the CLIP-Image encoder is resized to $224\times224$, and the model is finetuned for $100$ steps over the HICO-DET dataset, followed by $200$ steps over the GazeFollow dataset (batch size is $1024$ per step). The timestep sampling is piece-wise uniform, with the first $500$ timesteps twice as likely to be sampled as the last (noisier) $500$ timesteps. A constant learning rate of $5e-5$ is maintained for the UNet cross-attention layers and $5e-7$ for the Image-Text aligner layer. The inference is done with $30$ steps of UniPC~\cite{unipc_2023} sampler and guidance scale $5$. In all our experiments, the images are generated at a resolution of $512\times512$ unless otherwise specified.
\subsection{Evaluation metrics}
To evaluate the quality of the generated images, we compare the CLIP~\cite{pmlr-v139-radford21a} Text-Image alignment (\textbf{CLIP-T}) between the input prompts (captions) and the generated images. Additionally, we compare the CLIP Image-Image alignment (\textbf{CLIP-I}) and \textbf{DINO} score~\cite{dino_2021} between the reference images (or test set images for HICO-DET and GazeFollow) and the generated images using each of the baseline methods on each of the benchmark datasets. Besides the quality of images, we also want to evaluate the fidelity with respect to the human-object interactions, facial identities and the gaze patterns in the generated images. Notably, we are not constraining the layouts of the images at inference time, which can result in the reference subjects being generated at locations different from their ground-truth counterparts. Hence, for this evaluation, we first detect the interaction class labels, faces, objects and gaze targets in the generated images using appropriate detectors. Now, for interaction generation performance (on the generated HICO-DET test set), we calculate the mean Average Precision (\textbf{mAP}) of the CMMP~\cite{ting2024CMMP} detector across all the interaction classes, considering a label prediction as positive irrespective of the predicted bounding box location (i.e. no IoU constraint with the ground truth bounding boxes). For facial identity evaluation, we calculate the ArcFace~\cite{Deng_2019_CVPR} embeddings of each of the faces in each of the generated images, and then form (real, generated) identity embedding pairs for each image through greedy matching using cosine similarities. Afterwards, we calculate the \textbf{average cosine similarity} across all such pairs over all the test set images. Finally, for gaze pattern evaluation, we observe that in majority of the scenes, the humans are either looking at other humans or at some other object in the image. Therefore, we associate each of the gaze targets in the test set images with the object/human in whose bounding box it is present. The gaze of a head in a generated image is considered to have been rendered \textit{correctly} if the Sharingan~\cite{Tafasca_2024_CVPR}-detected gaze target lies within the bounding box of the same object label as in the corresponding ground truth image. Gaze targets which do not lie within any object's bounding box in the ground truth image are excluded from the evaluation. In this way, we obtain the \textbf{accuracy} of the gaze generation over the GazeFollow test set.

\subsection{Quantitative Results}
\begin{table}[t]
    \centering
    \begin{adjustbox}{max width=\textwidth}
    \begin{tabular}{l|c|c|c|c|c|c|c|c}
    \toprule
       \multirow{2}{*}{Method} &  \multirow{2}{*}{Arch}& \multicolumn{3}{c|}{mAP$(\uparrow)$}& Facial & \multirow{2}{*}{CLIP-T$(\uparrow)$}& \multirow{2}{*}{CLIP-I$(\uparrow)$}& \multirow{2}{*}{DINO$(\uparrow)$}\\
       \cline{3-5}
       & & Full & Rare & Non-rare & cos sim$(\uparrow)$& & &\\
    \midrule
    IP-Adapter~\cite{ye2023ipadaptertextcompatibleimage} &SDV1.5&13.22&13.44&13.15&0.5010&0.2969&0.7002&0.4579\\
    SSR-Encoder~\cite{Zhang_2024_CVPR} &SDV1.5&15.87&14.35&16.33&\textbf{0.6531}&\textbf{0.2971}&0.7231&0.5252\\
    SSR-Encoder*~\cite{Zhang_2024_CVPR} &SDV1.5&14.79&14.96&14.74&0.5715&0.2961&0.7217&0.5045\\
    \hline
    Ours &SDV1.5&\textbf{16.43}&\textbf{16.24}&\textbf{16.48}&0.6398&0.2966&\textbf{0.7241}&\textbf{0.5273}\\
    \hline
    Ours ($\mathcal{L}_{\text{boundary}}$)&SDV1.5&15.81&14.52&16.19&0.6505&0.2961&0.7226&0.5268\\
    Ours ($\mathcal{L}_{\text{id}}$)&SDV1.5&15.82&14.61&16.18&0.6440&0.2959&0.7241&0.5254\\
    Ours ($\mathcal{L}_{\text{gaze}}$)&SDV1.5&16.37&16.12&16.45&\textbf{0.6986}&\textbf{0.2986}&0.7197&\textbf{0.5325}\\
    Ours ($\mathcal{L}_{\text{pose}}$)&SDV1.5&16.20&14.86&16.60&0.6482&0.2964&\textbf{0.7251}&0.5280\\
    Ours ($\mathcal{L}_{\text{interaction}}$)&SDV1.5&\textbf{17.84}&\textbf{16.99}&\textbf{18.09}&0.5784&0.2930&0.7153&0.5216\\
    Ours (inverse timestep)&SDV1.5&14.09&12.55&14.55&0.5434&0.2910&0.7100&0.5204\\
    \hline
    MIP-Adapter~\cite{huang2024resolvingmulticonditionconfusionfinetuningfree} &SDXL&18.07&18.50&17.94&0.6358&0.3022&0.7412&0.5364\\
    \bottomrule
    \end{tabular}
    \end{adjustbox}
    \caption{Performance comparison on HICO-DET dataset. * indicates finetuning with only $\mathcal{L}_{\text{denoise}}$ and $\mathcal{L}_{\text{reg}}$. In rows 5-9, finetuning is done with the loss mentioned in Method column, along with $\mathcal{L}_{\text{denoise}}$ and $\mathcal{L}_{\text{reg}}$. Row 10 refers to finetuning with loss weights ($\lambda$) inversely proportional to the average values at different timesteps (shown in Figure 3.)}
    \label{tab:hicodet_results}
\end{table}
\begin{table}[t]
    \centering
    \begin{adjustbox}{max width=\textwidth}
    \begin{tabular}{l|c|c|c|c|c|c}
    \toprule
       \multirow{2}{*}{Method} & \multirow{2}{*}{Arch}&Gaze& Facial & \multirow{2}{*}{CLIP-T$(\uparrow)$}& \multirow{2}{*}{CLIP-I$(\uparrow)$}& \multirow{2}{*}{DINO$(\uparrow)$}\\
       &&Accuracy$(\%)(\uparrow)$&cos sim$(\uparrow)$&&\\
    \midrule
    IP-Adapter~\cite{ye2023ipadaptertextcompatibleimage}&SDV1.5&52.50&0.4786&\textbf{0.3054}&0.6797&0.4832\\
    SSR-Encoder~\cite{Zhang_2024_CVPR} &SDV1.5&53.81&0.6208&0.2966&0.6943&0.5202\\
    SSR-Encoder*~\cite{Zhang_2024_CVPR} &SDV1.5&52.46&0.5502&0.2976&0.6909&0.4917\\
    \hline
    Ours &SDV1.5&53.73&0.6237&0.2996&\textbf{0.6973}&0.5177\\
    Ours ($\mathcal{L}_{\text{gaze}}$)&SDV1.5&\textbf{56.39}&\textbf{0.6741}&0.3018&0.6968&\textbf{0.5237}\\
    \hline
    MIP-Adapter~\cite{huang2024resolvingmulticonditionconfusionfinetuningfree}&SDXL&57.68&0.6052&0.3022&0.7192&0.5503\\
    \bottomrule
    \end{tabular}
    \end{adjustbox}
    \caption{Performance comparison on GazeFollow dataset. * indicates finetuning with only $\mathcal{L}_{\text{denoise}}$ and $\mathcal{L}_{\text{reg}}$. In row 5, finetuning is done using $\mathcal{L}_{gaze}$ along with $\mathcal{L}_{\text{denoise}}$ and $\mathcal{L}_{\text{reg}}$.}
    \label{tab:gazefollow_results}
\end{table}
\begin{figure}[h]
    \centering
    \includegraphics[width=\linewidth]{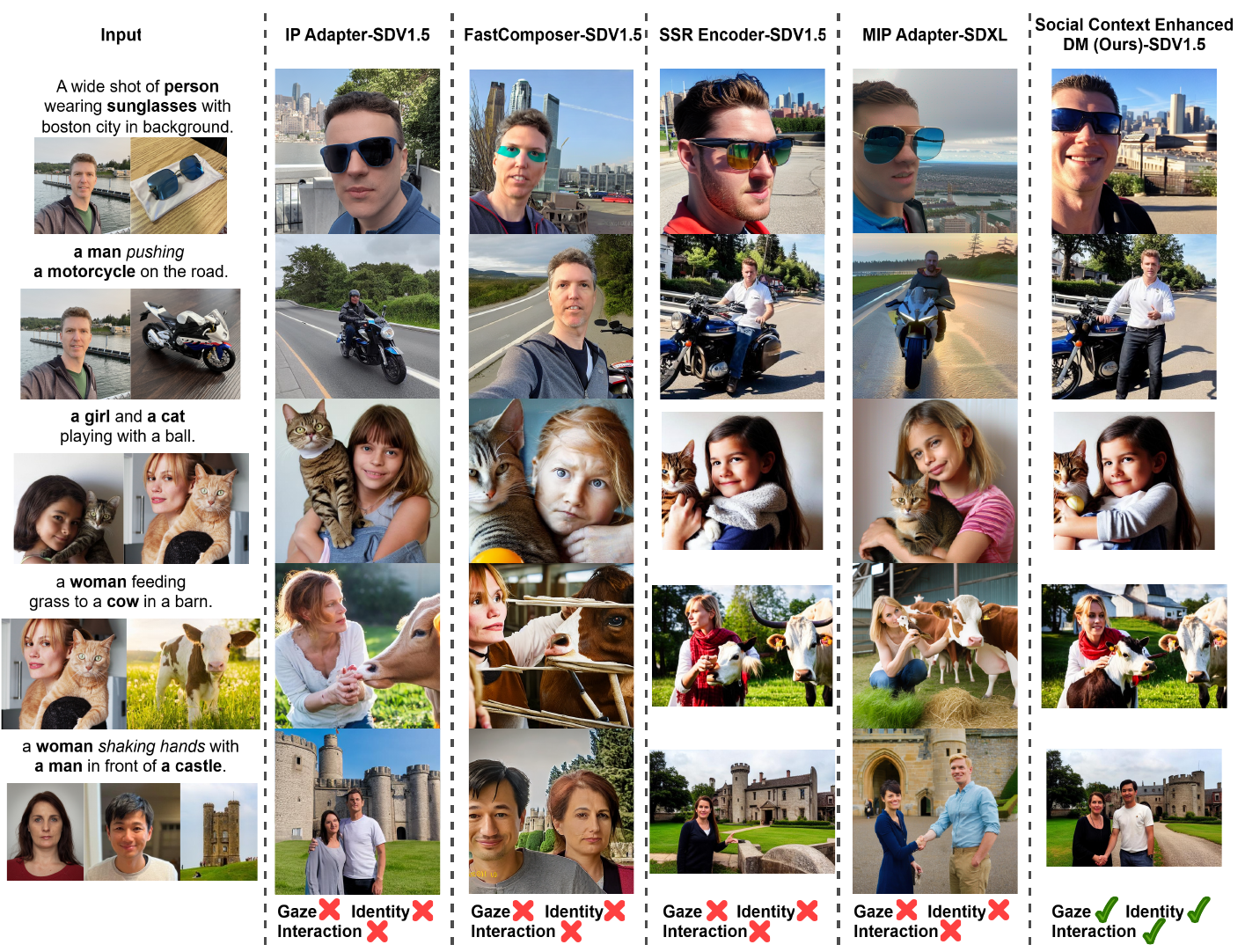}
    \caption{\textbf{Comparison with personalization baselines:} [Consider the first example as row 1, Input as column 1] In all five examples, prior methods exhibit inaccurate gaze (row 4: cols 2-4), identity (row 1: cols 3; row 2: cols 2, 5; row 3: cols 2, 3, 5; row 4: columns 2, 4; row 5: cols 2, 4, 5) and interaction (row 2: cols 2-5; row 3: cols 2, 4, 5; row 4: cols 2-5; row 5: cols 2-4). Our feedback-guided Diffusion model, leveraging pretrained gaze, identity, and interaction detectors, improves the generated images in each of these aspects.}
    \label{fig:comparison}
\end{figure}

In tables \ref{tab:hicodet_results}, \ref{tab:gazefollow_results} and \ref{tab:concept101_results}, we provide the quantitative comparison of our feedback based fine-tuning against several state-of-the-art methods on three different benchmarks. For interaction generation (Table \ref{tab:hicodet_results}), the mAP scores are improved across Rare (having $<10$ training images) as well as Non-rare interaction classes through feedback-based finetuning, when compared with SSR-Encoder. Though the facial embedding similarity and CLIP-Text scores are slightly reduced, the overall image quality is improved, as indicated by the better CLIP-I and DINO scores. The MIP Adapter method, being based on SDXL architecture (which is much larger than SDV1.5, and trained on larger amount of data), serves as the topline. Notably, its facial embedding similarity score is less than the SSR Encoder. IP-Adapter, on the other hand, having been trained with only single reference subjects, has lower scores across all the metrics. Also, the row 3 in the table indicates the performance of SSR-model when it is finetuned on the hicodet dataset using just the denoising and regularization losses (no social context feedback), which is worse than the original across all metrics (perhaps due to smaller training data size than original). For gaze generation (Table \ref{tab:gazefollow_results}), though our gaze accuracy is slightly reduced from the baseline SSR-Encoder, the facial embedding similarity, CLIP-Text and CLIP-Image similarity are improved. Row 3 indicates a loss of performance upon finetuning without social context feedback for this dataset as well. Similar to HICO-DET performance results, MIP Adapter, with a larger base model, has better scores across all the metrics except facial embedding similarity.
\begin{wraptable}{i}{0.6\textwidth}
\vspace{-\baselineskip}
    \centering
    \begin{adjustbox}{max width=0.6\textwidth}
    \begin{tabular}{l|c|c|c}
    \toprule
       Method & CLIP-T$(\uparrow)$& CLIP-I$(\uparrow)$& DINO$(\uparrow)$\\
    \midrule
    IP-Adapter~\cite{ye2023ipadaptertextcompatibleimage}&0.6343&0.6409&0.3481\\    FastComposer~\cite{osti_10543865}&0.7456&0.6552&0.3574\\
    SSR-Encoder~\cite{Zhang_2024_CVPR} &\textbf{0.7567}&0.6728&0.3694\\
    MIP-Adapter~\cite{huang2024resolvingmulticonditionconfusionfinetuningfree}& 0.7410& 0.6704&0.3350\\
    \hline
    Ours &0.7548&\textbf{0.6738}&\textbf{0.3704}\\
    \bottomrule
    \end{tabular}
    \end{adjustbox}
    \caption{Performance comparison on Concept101.}
    \label{tab:concept101_results}
    \vspace{-\baselineskip}
\end{wraptable}
We also present a comparison on the popular Concept101 benchmark, for image quality assessment in personalized generation. As Table \ref{tab:concept101_results} demonstrates, our finetuning achieves better CLIP-Image and DINO scores across all the methods (even SDXL based MIP Adapter, sampled at $512\times512$ resolution).\\
\noindent\textbf{Ablation:} To reveal the impact of each of the losses individually, we perform an ablation study, whose results are mentioned in Table \ref{tab:hicodet_results} (rows 5-10) and Table \ref{tab:gazefollow_results} (row 5). In each row, the model has been finetuned using the loss indicated in the method column, in addition to the denoising and regularization losses. The interaction-based feedback alone provides the best interaction generation (with the best mAP scores), though at the cost of reduced facial identity and semantic alignments. Interestingly, gaze-based feedback results in a much better facial identity similarity as compared to identity based feedback. Using gaze feedback alone provides considerably improved gaze generation accuracy, facial identity similarity and image quality (DINO scores). An additional experiment to show the effect of a different weighing strategy ($\lambda$ inversely proportional to the loss curves in Figure \ref{fig:tstep_loss} (left)) results in a severe performance degradation, thus demonstrating the superiority of the step-based supervision chosen in Section \ref{sec:Method}(F).
\subsection{Qualitative Results}
Qualitatively, we show the feedback-based improvement in the generation of identity, gaze and interaction in Figure \ref{fig:comparison}. In example 1 (topmost), while FastComposer and IP Adapter are unable to generate the sunglasses properly, the person generated by SSR-Encoder seems like a doppelganger of the input person. In the second example, all the existing methods fail to generate a man \textit{pushing} a motorcycle; rather they either totally occlude the motorcycle (FastComposer), or show the man $riding$ the motorcycle. In example 3, the ball (whose reference image is not provided) is not generated by IP-Adapter, SSR-Encoder and MIP-Adapter. The generated girl's identity is also dissimilar from the reference image for IP-Adapter, FastComposer and MIP-Adapter. In example 4, the cow is improperly generated by IP-Adapter and SSR-Encoder, and none of the existing methods are able to properly generate the \textit{feeding} interaction between the woman and the cow. In IP-Adapter, the woman's hands are folded together, while FastComposer, SSR-Encoder and MIP-Adapter show the woman as looking away, when she should be looking \textit{towards} the cow to feed her. In the last example, the identities of the generated man and woman are dissimilar from the inputs for IP-Adapter and MIP-Adapter, and SSR-Encoder fails to generate the man altogether. In IP-Adapter, the man puts his hand on the shoulder of the woman, instead of shaking hands, and in FastComposer, there is no interaction between the man and the woman. In all these examples, feedback based finetuning leads to proper generation of identity, gaze as well as interaction.
\section{Conclusion}
In this work, we propose a way to improve the quality of personalized image generation \textit{in social context} by carefully incorporating multiple feedback signals from pretrained detectors into the diffusion model's finetuning. Depending upon the granularity of the feedback signal, we also adjust the timestep ranges where the feedback is utilized. Qualitative and quantitative experiments demonstrate the effectiveness of our approach in improving gaze, identity and interaction generation for personalized images, while maintaining superior image quality.

\bibliography{egbib}
\end{document}